\documentclass[11pt,journal,compsoc]{IEEEtran}
\ifCLASSOPTIONcompsoc
  \usepackage[nocompress]{cite}
\else
  \usepackage{cite}
\fi
\usepackage{threeparttable}

\ifCLASSINFOpdf
   \usepackage[pdftex]{graphicx}
\else
\fi
\usepackage{url}
\usepackage{longtable}

\begin{document}
%
\title{Transformer-based Korean Pretrained Language Models: A Survey on Three Years of Progress}
%
%
%
%

\author{Kichang~Yang,~\IEEEmembership{Undergraduate Student,~Soongsil University}
\thanks{contact: ykcha9@gmail.com}}

\IEEEtitleabstractindextext{%
\begin{abstract}
With the advent of Transformer, which was used in translation models in 2017, attention-based architectures began to attract attention. Furthermore, after the emergence of BERT, which strengthened the NLU-specific encoder part, which is a part of the Transformer, and the GPT architecture, which strengthened the NLG-specific decoder part, various methodologies, data, and models for learning the Pretrained Language Model began to appear. Furthermore, in the past three years, various Pretrained Language Models specialized for Korean have appeared. In this paper, we intend to numerically and qualitatively compare and analyze various Korean PLMs released to the public.
\end{abstract}

}

\maketitle

\IEEEdisplaynontitleabstractindextext

%
\IEEEpeerreviewmaketitle

\IEEEraisesectionheading{\section{Introduction}\label{sec:introduction}}

%
%
%
%
\IEEEPARstart{T}{he} 
hot keyword in the field of Natural Language Processing and, furthermore, Machine Learning for the last three years, was the Transformer\cite{vaswani2017attention}-based BERT\cite{devlin2019bert} or GPT\cite{radfordimproving} model using the attention algorithm. Originally, the Transformer was a model focused on the NMT model that appeared because of the gradient bottleneck problem that occurred during train RNN\cite{saklong} for implementing the Neural Machine Translation model, but BERT, which performs only the role corresponding to NLU in the Transformer, and GPT, which performs only the role corresponding to NLG. With the advent of the two models, various models, algorithms, and data pre-processing methods appeared. In addition to this, as a pretrained-model sharing platform called "Transformers"\cite{wolf2019huggingface} created by huggingface\footnote{\url{https://huggingface.co/}} appeared, the NLP/AI field achieved unprecedented growth in both academia and industry. 
Most recently, large-scale models such as GPT3\cite{brown2020language} that scale-up the parameters and data of GPT by hundreds of times or more, or models that expanded (ViT\cite{dosovitskiy2020image}) or mixed (DALL-E\cite{DBLP:journals/corr/abs-2102-12092}) modality was appeared, it seemed to be getting a little closer to AGI.
On the other hand, based on this huggingface platform and the Transformer series model, active research and development of models specialized in the Korean domain were also conducted in many areas of companies, schools, and individuals.
Accordingly, we would like to conduct a comprehensive survey by combining the research results of individual researchers/developers  and Korean companies such as Naver\footnote{\url{https://www.navercorp.com}}, Kakao\footnote{\url{https://www.kakaocorp.com}}, and SKT\footnote{\url{https://www.sktelecom.com}} do.
In this paper, the contribution we would like to claim is as follows.

\begin{itemize}
\setlength\itemsep{.2em}
\item Introduction and summary of the type of Korean models that have been released so far
\item Introduction and arrangement of Korean benchmark datasets that have been released so far
\item Comprehensive score analysis of published models.
\end{itemize}

\begin{figure} [h]
   \centering
   \includegraphics[width=\columnwidth]{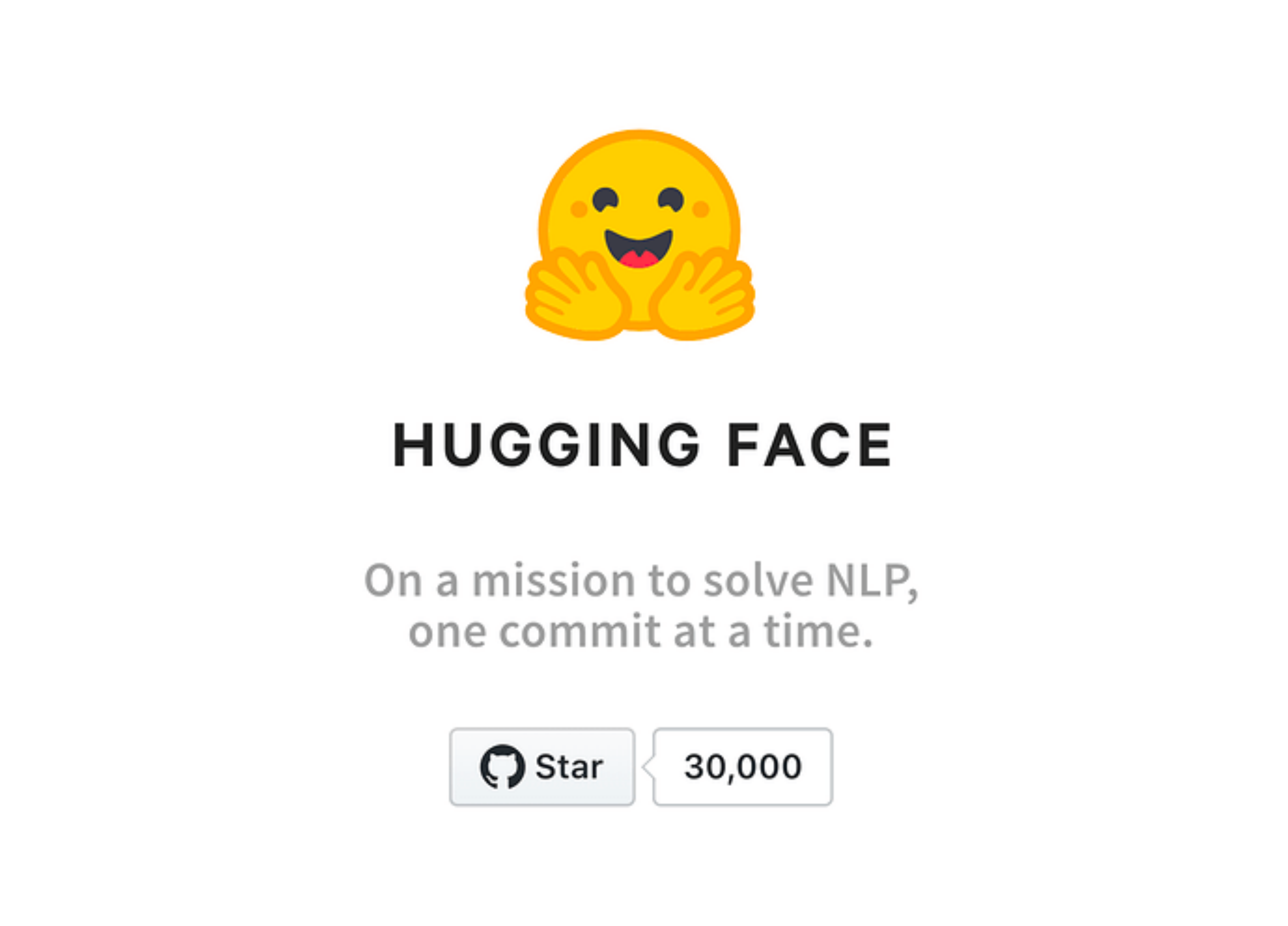}
   \caption{The emergence of hugging face provided a platform to share the pretrained model in the natural language processing field and machine learning field, and at the same time brought the revival of the Transformer model-based attention algorithm, leading to tremendous growth in both the industry / academia in the AI and NLP fields.} 
   \label{fig:figure0}
\end{figure} 

\section{Related Works}
\subsection{Neural Machine Translation}
The most popular framework for NMT is the encoder-decoder model\cite{cho2014learning,sutskever2014sequence,bahdanau2014neural,luong2015effective,vaswani2017attention}. Adopting attention module greatly improved the performance of encoder-decoder model by using context vector instead of fixed length vector\cite{bahdanau2014neural, luong2015effective}. By exploiting multiple attentive heads, the Transformer model has become the de-facto standard model in NMT\cite{vaswani2017attention,  ott2018scaling,so2019evolved}.

\subsection{Pretraining with Unsupervised Feature-based Approaches}
In recently, there are several approaches for pretraining methods of the main stream, using feature-based approaches. 
OpenAI GPT\cite{radfordimproving} uses decoder of Transformer architecture with next token prediction (Auto Regressive) object. 
On the other side of GPT, BERT\cite{devlin2019bert} uses submodule of Transformer architecture(encoder), with Masked Language Modeling(MLM) and Next Sentence Prediction(NSP) object for pretraining. RoBERTa\cite{liu2019roberta} is similar to BERT architecture except it trained without NSP object and static masking during the pretraining process. ELECTRA\cite{clark2020electra} uses MLM object with an adversarial objective used in GAN\cite{goodfellow2020generative} architecture for pretraining, using only a discriminator in finetuning different from GAN. 
BART\cite{lewisbart} uses both an encoder and decoder (ie. full architecture of transformer) architecture with several permutation and deletion objectives.
In the finetuning step, simply plugging in the task-specific inputs and outputs into each PLM, we introduce and finetune all of the parameters end-to-end. 
However, recent researches of Large-Scale PLM like GPT-3\cite{brown2020language} show that any finetuning steps are not needed as the size of PLM and data are large enough to remember all of tasks and information from training data. However, as there are very few number of large-scale Korean PLMs exist, our survey does not include these types of PLMs.

\begin{figure*}
   \centering
   \includegraphics[scale=0.9]{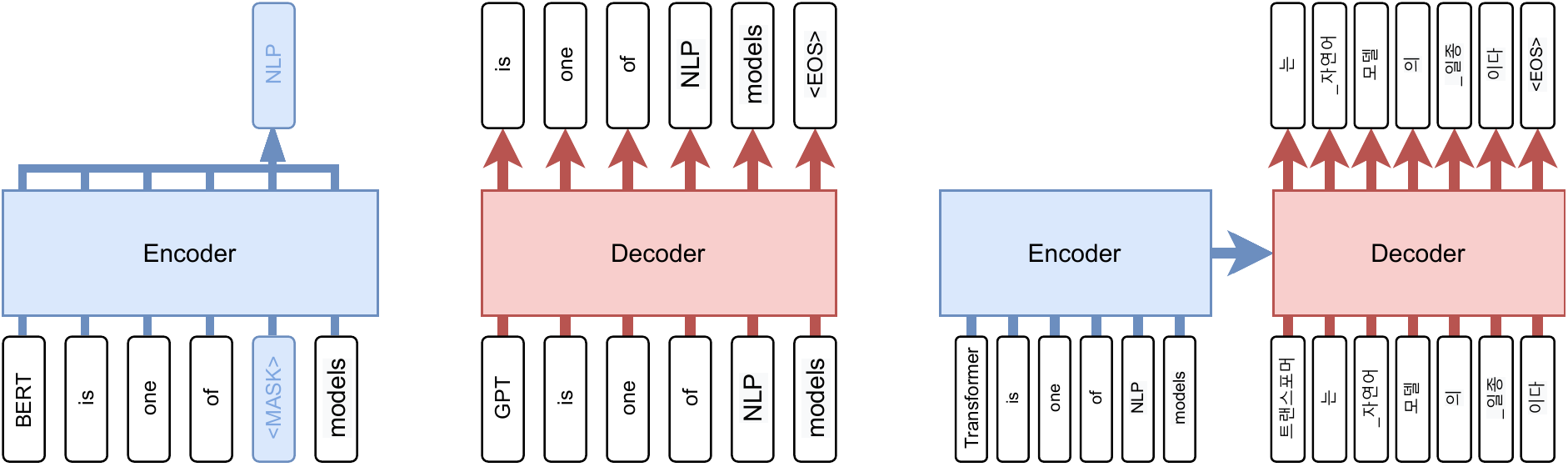}
   \caption{
   Three main types of PLM. Encoder-Centric Models (left) trained with MLM Task for Language Understanding, Decoder-Centric Models(center) trained with Next Token Prediction Task for Language Generation. Seq2Seq Models (right) trained with various objects and tasks (NMT, Summarization, etc) with Next Token Prediction for language generation with understanding. Detailed objectives or architectures can be differ by individual models. } 
   \label{fig:figure1}
\end{figure*} 
\subsection{Korean NLP Benchmarks}
Various fine-tuning data and test sets that can measure the performance of Korean natural language tasks have been released. NSMC\footnote{https://github.com/e9t/nsmc} dataset is a Sentiment Analysis dataset labeled on Naver movie review comment data. Naver and Changwon University unveiled NaverNER\footnote{https://github.com/naver/nlp-challenge}, the Korean NER data, at a competition held together in 2018. Kakao Brain released the KorNLI and KorSTS\cite{DBLP:journals/corr/abs-2004-03289} datasets for measuring the NLU performance of Korean in 2020. Also, in 2019, LG CNS released KorQuAD\footnote{https://korquad.github.io/}, a SQuAD dataset for Korean to measure the performance of Korean question and answer tasks. In 2020, the BEEP!\cite{moon-etal-2020-beep} dataset for the Korean Hate Speech Classification task was released. Most recently, KLUE\cite{park2021klue}, Korean version of GLUE\cite{wang2019glue} benchmark was released. However, we are not reporting this results as lots of models are not reported for this benchmark yet.

\section{Korean PLM Architectures}
Language models after 2018 can be classified into three major types according to the pretraining method (Fig. \ref{fig:figure1}). (1) The first group of models is Encoder-Centric Models, which focus on “Understanding” of language (NLU) by using objective functions such as predicting the corresponding MASK (MLM) after creating and inserting the MASK in the input sentence. These models are later fintuned for tasks such as classification or feature extraction. As a representative model, BERT series PLMs are applicable. (2) The second case is using the objective function to predict the next token of each input token. Since these models are optimized for inference corresponding to Auto-Regressive, they are mainly used for Downstream-Task (Chat-bot, Lyric Generation, etc) learning corresponding to Language Generation (NLG). This mainly applies to GPT-based PLMs. (3) The third case is a model that utilizes the entire architecture of Transformer, which has recently been introduced in many ways. Models such as T5\cite{raffel2020exploring}, BART\cite{lewisbart}, and MASS\cite{song2019mass} are representative. The model trained in this way shows some significant performance improvement not only in NLU and NLG, but also in tasks where the effect of PLM is hard to see, such as NMT. In this section, we introduce the tokenizers and parameters of the Korean pretrained models that have been released so far based on the three categories we classified above.

\subsection{Encoder-Centric Models}
\begin{figure} [h]
   \centering
   \includegraphics[width=\columnwidth]{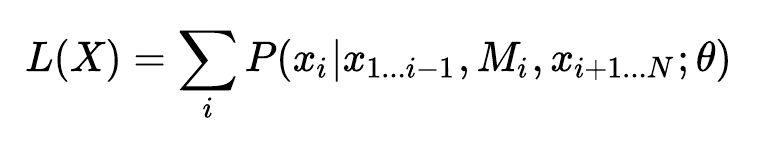}
   \caption{Objective of Encoder-Centric Models. this objective is to predict tokens that positions of input are masked, designed for Natural Language Understanding.} 
   \label{fig:figure2}
\end{figure} 

Encoder centric models are focused on extracting the features of language. Some tasks like classification, clustering, tagging are able to use this type of models as PLM.

\subsubsection{KoBERT} KoBERT\footnote{\url{https://github.com/SKTBrain/KoBERT}} is the first Korean pretrained model shared on huggingface released by SKT-Brain. It is mostly the same as BERT's configuration, but the tokenizer uses SentencePiece \footnote{\url{https://github.com/google/sentencepiece}}, not the Word Piece Toeknizer used in BERT. For the data used for pretraining, 5 million sentences and 54 million words were used in the Korean wiki.

\subsubsection{HanBERT} HanBERT\footnote{\url{https://github.com/tbai2019/HanBert-54k-N}} is a BERT model trained using about 150GB (General Domain: 70GB, Patent Domain: 75GB) and 700 million sentences of Korean corpus. The tokenizer used a private tokenizer called Moran Tokenizer, and the vocab size was 54000.

\subsubsection{KoELECTRA}
KoELECTRA\footnote{\url{https://github.com/monologg/KoELECTRA}} is an ELECTRA-based language model trained from `Modu Corpus'\footnote{\url{https://corpus.korean.go.kr/}} released by the National Institute of Korean Language (NIKL), Korean Wikipedia, NamuWiki\footnote{Large-scale Korean open domain encyclopedia.}, and various news data.

\subsubsection{KcBERT}
KcBERT\cite{lee2020kcbert} is a Korean BERT model trained based on the BERT model using about 12 GB of Naver politics news comments data. Tokenizer uses Wordpiece\cite{schuster2012japanese}  BPE and is preprocessed to handle emojis and special characters.

\subsubsection{SoongsilBERT (KcBERT2)}
SoongsilBERT\footnote{\url{https://github.com/jason9693/Soongsil-BERT}} is a language model pretrained by using community data of Soongsil University and Modu Corpus in addition to the news comments data used in KcBERT. Most of the settings are identical, except it is trained based on the RoBERTa model and uses Byte-level BPE Tokenizer. Moreover, SoongsilBERT is more fitting to community terminology. In other words, it does not perform well in non-community domains. 

\subsubsection{KcELECTRA}
KcELECTRA\footnote{\url{https://github.com/Beomi/KcELECTRA}} is a model trained by collecting additional data (mainly comments) to the data used for KcBERT. In NSMC Task, the model is currently recording State-of-the-Arts.

\subsubsection{DistilKoBERT}
DistilKoBERT\footnote{\url{https://github.com/monologg/DistilKoBERT}} is a lightweighted version of KoBERT distillation based on huggingface's DistilBert\cite{sanh2019distilbert}model. The Teacher model and tokenizer used are the same as KoBERT.

\subsubsection{KoBigBird}
KoBigBird\cite{jangwon_park_2021_5654154} is released for long-range understanding of Korean language. This model covered with more than 8 times longer than the usual (512 tokens) BERT models. 
 
\subsection{Decoder-Centric Models}
\begin{figure} [h]
   \centering
   \includegraphics[width=\columnwidth]{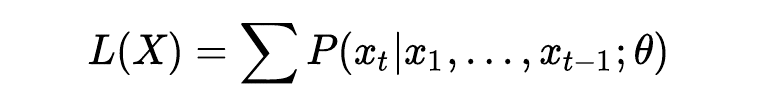}
   \caption{Objective of Decoder-Centric Models. this objective is just predict next token of each position of input tokens, designed for Auto-Regressive inference, which forward model iterative until eos (end-of-sequence) token predicted.} 
   \label{fig:figure3}
\end{figure}
Decoder centric models are focused on a generation of languages. Some tasks like dialog (someone called chatbot), Lyric-generation, or other types of "generate" language are able to use this type of models as PLM. Like fig. \ref{fig:figure3}, objective functions of Decoder-centric models are too simple, just predicting the next token of all sequences is done. 
Unfortunately, in Korean, a very few number of models that trained on this type are released as many PLM focused on NLU, not NLG.

\subsubsection{SKT-AI KoGPT2}
KoGPT2\footnote{\url{https://github.com/SKT-AI/KoGPT2}} is a language model that GPT2\cite{radford2019language}-based PLM for the first Korean Natural Language Generation released by SKT-AI. Korean Wikipedia, Modu Corpus, and the Blue House National Petition\footnote{\url{https://github.com/akngs/petitions}} and private data like news were used for model training. Char BPE Tokenizer is used for tokenization, in addition to the custom (unused) tokens that used to train the downstream task.

\subsubsection{Large-Scale PLM}
As mentioned above, in abstraction, we will introduce about large-scale LMs for Korean, but we will not deal with these models because of computational limitations.
\begin{itemize}
\setlength\itemsep{.2em}
     \item \textbf{HyperCLOVA\cite{kim2021changes}}: HyperCLOVA is first version of Korean Large-Scale PLM. Parameter size is up to 82B, but the models(i.e parameters) are not published now.
     \item \textbf{SKT KoGPT-trinity}: SKT KoGPT-trinity (We'll call this model as \textbf{SKGPT}) is the first public version of Large-scaled Korean PLM. The size of parameters is 1.2B and trained with Ko-DATA dataset which is inner refined corpus of SKT for training the model.
     \item \textbf{KakaoBrain KoGPT\cite{kakaobrain2021kogpt}\footnote{\url{https://github.com/kakaobrain/kogpt}}}: Kakao Brain's KoGPT(for preventing confused with KoGPT2 released by SKT, We'll call this model as \textbf{KakaoGPT}) is the largest(size of model) public version of Korean PLM. Parameter size is 6B.
\end{itemize}
SKGPT and KakaoGPT announced with their down-stream task results as finetuning, unlike HyperCLOVA registered as prompt-tuning version.

\subsection{Seq2Seq-Centric Models}
\begin{figure} [h]
   \centering
   \includegraphics[width=\columnwidth]{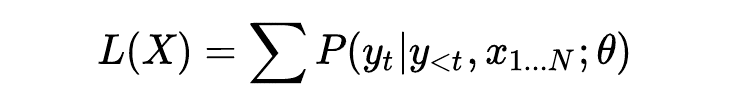}
   \caption{Objective of Seq2Seq-Centric Models. this type's objective is usually mixed with NLU and NLG function.} 
   \label{fig:figure3}
\end{figure}
Seq2Seq\cite{sutskever2014sequence} centric models use seq2seq transformer architecture for both NLU and NLG. Lots of pretraining method are available as so many seq2seq tasks exist. Unfortunately, in Korean, a few numbers of PLM trained with this method are opened.
\subsubsection{KoBART}
KoBART\footnote{\url{https://github.com/SKT-AI/KoBART}} is one of Seq2Seq versions of PLM, based on BART model, which have training objects: text infilling (for NLU) and Auto-Regressive(for NLG). It was trained on more than 40GB corpus.

\section{Experiment}
\begin{table*}[]
\renewcommand{\arraystretch}{1.3}
\caption{
A Result of Single Sentence Tasks.
}
\centering
\label{result}
\begin{threeparttable}
\begin{tabular}{|l|c|c|c||c|}
\hline
\textbf{Models}     & \textbf{NSMC}\tnote{*} & \textbf{BEEP!(Dev)}\tnote{2} & \textbf{Naver NER}\tnote{3} & \textbf{Size(MB)} \\ \hline
KoELECTRA(Small)    & 89.36         & 63.07               & 85.4               & 54       \\
KoELECTRA(Base)     & 90.63         & 67.61               & \textbf{88.11}     & 431      \\
DistilKoBERT        & 88.6          & 60.72               & 84.65              & 108      \\
KoBERT              & 89.59         & 66.21               & 87.92              & 351      \\
SoongsilBERT(Small) & 90.7          & 66                  & 84                 & 213      \\
SoongsilBERT(Base)  & 91.2          & 69                  & 85.2               & 370      \\
KcBERT(Base)        & 89.62         & 68.78               & 84.34              & 417      \\
KcBERT(Large)       & 90.68         & 69.91      & 85.53              & 1200     \\
KoBigBird(Base)     & 91.18         & -                   & -                  & 436      \\
KoBART              & 90.24         & -                   & -                  & 473      \\
KoGPT2              & 91.13         & -                   & -                  & 490      \\
HanBERT             & 90.06         & 68.32               & 87.70              & 614      \\
XLM-Roberta-Base    & 89.03         & 64.06               & 86.65              & 1030     \\
KcELECTRA-base	    & \textbf{91.71}	        & \textbf{74.05}                   & 86.90	           & 475      \\
\hline
\end{tabular}
\begin{tablenotes}\footnotesize
\item[1] measured by accuracy.
\item[2, 3] measured by F1 score.
\end{tablenotes}
\end{threeparttable}

\end{table*}
We exploit an aggregate Down-Stream task benchmark results of several pretrained models we discussed above. Using the benchmark datasets introduced in related works, We report results with 2 aspects. (1) Tasks deal with only a single sentence task (TABLE ~\ref{result}), and (2) Tasks deal with multiple sentences or have some interactions with multiple agents (TABLE ~\ref{table2}).
\subsection{Single Sentence Tasks}
Korean Benchmarks with a single sentence are mainly focused on classification or tagging task. NSMC is Korean sentiment classification benchmark which have binary classes, labeled with NAVER Corp's Shopping review comments. and BEEP! is Korean hate-speech classification benchmark labeled with "Hate", "Offensive" and "None" classes. Naver NER is namely Named Entity Recognition benchmark of Korean, which opend by NAVER Corp.

\subsubsection{NSMC Result}
NSMC is one of the benchmark datasets classifying whether sentiment is positive or negative. All sentences are come from the commercial review sentence of NAVER. The size of this dataset is 150k sentences for training and 50k sentences for testing. KcELECTRA has recorded State-of-the-Art (SOTA) in this task with 91.71 accuracy. 

\subsubsection{BEEP! Result}
BEEP! is a human-annotated corpus where the intensity of hate speech is tagged with the labels of `hate', `offensive', and `none',  built upon celebrity news comments on a Korean online news platform. KcELECTRA achieved highest score of the models with 69.91 F1 Scores. One of the interesting things is that DistilKoBERT, lightweight version of KoBERT, the score is degraded more than 5 points unlike NSMC scores of the two models are nearly the same. 

\subsubsection{Naver NER Result}
Naver NER dataset is a data published by processing Korean Wikipedia into text form. Total number of training sets is 90,000 examples. KoELECTRA (Base) model achieved State-of-the-Art in this task. One of the interesting things is KcBERT and SoongsilBERT, unlike in NSMC or BEEP! results, these models are not performed well, even worse than the general multilingual model (XLM)\cite{DBLP:journals/corr/abs-1911-02116} that is not specialized in Korean. 

\begin{table*}[!t]
\renewcommand{\arraystretch}{1.3}
\caption{A Result of Multiple Sentence \& Agent Tasks.}
\label{table2}
\centering
\begin{threeparttable}
\begin{tabular}{|l|c|c|c|c||c|}
\hline
\textbf{Models}   & \textbf{KorNLI}\tnote{1} & \textbf{KorSTS}\tnote{2} & \textbf{Question Pair}\tnote{3} & \textbf{KorQuaD (Dev)}\tnote{4} & \textbf{Size (MB)}  \\ \hline
KoELECTRA (Small)    & 78.6            & 80.79           & 94.85                  & 82.11 / 91.13         & 54   \\
KoELECTRA (Base)     & \textbf{82.24}           & \textbf{85.53}                  & 95.25                 & 84.83 / 93.45         & 431  \\
DistilKoBERT        & 72              & 72.59           & 92.48                  & 54.40 / 77.97         & 108  \\
KoBERT              & 79.62           & 81.59           & 94.85                  & 51.75 / 79.15         & 351  \\
SoongsilBERT (Small) & 76              & 74.2            & 92                     & -                     & 213  \\
SoongsilBERT (Base)  & 78.3            & 76              & 94                     & -                     & 370  \\
KcBERT (Base)        & 74.85           & 75.57           & 93.93                  & 60.25 / 84.39         & 417  \\
KcBERT (Large)       & 76.99           & 77.49           & 94.06                  & 62.16 / 86.64         & 1200  \\
KoBigBird (Base)     & -               & -               & -                      & \textbf{87.08 / 94.71}         & 436  \\
KoBART              & -               & 81.66           & 94.34                  & -                     & 473  \\
KoGPT2              & -               & 78.4            & -                      & -                     & 490  \\
HanBERT             & 80.32           & 82.73           & 94.72                  & 78.74 / 92.02         & 614  \\
XLM-Roberta-Base    & 80.23           & 78.45           & 93.8                   & 64.70 / 88.94         & 1030 \\
KcELECTRA-base	    & 81.65           & 82.65           & \textbf{95.78}                  & 70.60 / 90.11         & 475  \\
\hline
\end{tabular}
\begin{tablenotes}\footnotesize
\item[1, 3] measured by accuracy.
\item[2] measured by spearman correlation.
\item[4] measured by (1) EM score and (2) F1 score.
\end{tablenotes}
\end{threeparttable}
\end{table*}

\subsection{Multiple Sentence and Agent Tasks}
The result of this task showed different patterns than before. KoELECTRA and KoBigBird showed best results, whereas KcBERT and SoongsilBERT showed better before. In this task, unlike before, the texts of the datasets are much longer as there are some interaction between sentences (NLI, STS) or agents (QA). KorNLI and KorSTS are NLI dataset for Korean, released by Kakao Brain. and Question Pair (Korean) dataset is a pharaphrase detection benchmark that finds the similarity between two question sentences for Korean. Unfortunately, we can not access this dataset anymore as this repo is removed now. Finally, KorQuAD dataset is a Korean version of SQuAD (QA) dataset. Although the latest version of this dataset is 2.0, We used 1.0 as lots of models reported in this version.

\subsubsection{KorNLI Result}
NLI task is the task classifying the relationship between two sentences as "entailment", "contradiction" and "neutral". KorNLI dataset has 942,854 examples (pair) for training, 2,490 exmamples for evaluation, and 5,010 examples for testing. KoELECTRA scored State-of-the-Arts in this task. However, most of Korean PLM scored lower points than XLM, not Korean centered model.

\subsubsection{KorSTS Result}
STS task is identical to NLI, except for scoring metric. This dataset score similarity between two sentences from 1 (not similar) to 5 (identical). In KorSTS, it has 5,749 examples for training, and 1,500 examples for evaluation, and 1,379 examples for testing. Like 4.2.1, KoELECTRA recorded the best score in this task.

\subsubsection{Question Pair Result}
Question Pair dataset has 6,888 examples of train sets and 688 examples of test sets. KcELECTRA model has recorded the best in this task. However, 
Question Pair dataset is currently unavailable because the repository of this task and dataset is vanished.

\subsubsection{KorQuAD Result}
The total data of KorQuAD are divided into 10,645 paragraphs and 66,181 Q\&A pairs for 1,560 Wikipedia articles, 60,407 Q\&A pairs for the training set, and 5,774 Q\&A pairs for Dev set. In this task, KoBigBird scored the highest score (87.08 EM score / 94.71 F1-score), On the other hand, KcBERT and KoBERT are not performed well even scored lower than XLM. It seems the sentence length of corpus for pretraining is too short to understand long-term sequence.

\section{Conclusion}
In this survey, we discussed several Korean pretrained language models and benchmarks and compared with these models. Of course, there are many more publicly available Korean language models other than the model we introduced, but we could not include all of them for reasons such as the length of this paper or the reason that the benchmark results were not reported in various ways. In future works, we expect that the latest Korean benchmarks such as KLUE and various surveys will appear to promote the development of Korean NLP and furthermore, Computational Linguistics.


%

\appendices




\ifCLASSOPTIONcaptionsoff
  \newpage
\fi



\bibliographystyle{IEEEtran}
\bibliography{IEEEexample}
\end{document}